\pdfoutput=1

\documentclass[11pt]{article}

\usepackage{naacl2021}

\usepackage{times}
\usepackage{latexsym}
\usepackage{booktabs} 

\usepackage[T1]{fontenc}

\usepackage[utf8]{inputenc}

\usepackage[cmex10]{amsmath}
\usepackage{amsfonts, amsthm}
\usepackage{algorithm}
\usepackage{algpseudocode}

\usepackage{tikz}
\usetikzlibrary{arrows, decorations.pathreplacing, decorations.pathmorphing,backgrounds,positioning,fit,petri}

\newtheorem{theo}{Theorem}
\newtheorem{lemm}[theo]{Lemma}

\newtheorem{prop}[theo]{Proposition}

\theoremstyle{remark}

\usepackage{microtype}

\def\A {\mathbf{A}}

\def\S {\mathbf{S}}

\def\X {\mathbf{X}}

\def\s {\boldsymbol{s}}

\def\x {\boldsymbol{x}}
\def\y {\boldsymbol{y}}

\def\Cc {\mathcal{C}}

\def\Ic {\mathcal{I}}

\def\Lc {\mathcal{L}}

\def\Pc {\mathcal{P}}

\def\Sc {\mathcal{S}}
\def\Tc {\mathcal{T}}

\def\Rb {\mathbb{R}}
\def\Pb {\mathbb{P}}
\def\Eb {\mathbb{E}}

\newcommand{\argmax}[1]{\underset{#1}{\operatorname{argmax\,\,}}}

\usepackage[retainorgcmds]{IEEEtrantools}
\newcommand{\be}{\begin{IEEEeqnarray*}{rCl}}
\newcommand{\ee}{\end{IEEEeqnarray*}}

\newcommand{\ben}{\begin{IEEEeqnarray}{rCl}}
\newcommand{\een}{\end{IEEEeqnarray}}

\title{Masked Conditional Random Fields for Sequence Labeling}

\author{Tianwen Wei\thanks{\quad Corresponding author.}, Jianwei Qi, Shenghuan He, Songtao Sun \\
Xiaomi AI \\
\texttt{\{weitianwen,qijianwei,heshenghuan,sunsongtao\}@xiaomi.com}}

\begin{document}
\maketitle
\begin{abstract}
Conditional Random Field (CRF) based neural models are among the most performant methods for solving sequence labeling problems. Despite its great success, CRF has the shortcoming of occasionally generating illegal sequences of tags, e.g. sequences containing an ``I-'' tag immediately after an ``O'' tag, which is forbidden by the underlying BIO tagging scheme. In this work, we propose Masked Conditional Random Field (MCRF), an easy to implement variant of CRF that impose restrictions on candidate paths during both training and decoding phases. We show that the proposed method thoroughly resolves this issue and brings consistent improvement over existing CRF-based models with near zero additional cost. 
\end{abstract}

\section{Introduction}
Sequence labeling problems such as named entity recognition (NER), part of speech (POS) tagging and chunking have long been considered as fundamental NLP tasks and drawn researcher's attention for many years. 

Traditional work is based on statistical approaches such as Hidden Markov Models \citep{HMM} and Conditional Random Fields \citep{FIRST_CRF}, where handcrafted features and task-specific resources are used. With advances in deep learning, neural network based models have achieved dominance in sequence labeling tasks in an end-to-end manner. Those models typically consist of a neural encoder that maps the input tokens to embeddings capturing global sequence information, and a CRF layer that models dependencies between neighboring labels. Popular choices of neural encoder have been convolutional neural network \citep{CNN_CRF}, and bidirectional LSTM \citep{huang_bidirectional_2015}.
Recently, pretrained language models such as ELMo \citep{peters_deep_2018}
or BERT \citep{BERT} have been proven far superior as a sequence encoder, achieving state-of-the-art results on a broad range of sequence labeling tasks.


Most sequence labeling models adopt a BIO or BIOES tag encoding scheme \citep{RATINOV2009}, which forbids certain tag transitions by design. Occasionally, a model may yield  sequence of predicted tags that violates the rules of the scheme. Such predictions, subsequently referred to as \emph{illegal paths}, are erroneous and must be dealt with. 
Existing methods rely on hand-crafted post-processing procedure to resolve this problem, typically by retaining the illegal segments and re-tagging them. But as we shall show in this work, such treatment is arbitrary and leads to suboptimal performance.

\begin{table*}[th]
\begin{center}
\begin{tabular}{r|c|c|c|c|c|c|c}
\toprule
 Dataset  & {\small legal \& TP} & {\small illegal \& TP} & {\small legal \& FP} & {\small illegal \& FP} 
 & $\frac{\textrm{illegal \& TP}}{\textrm{illegal}}$  
 & $\frac{\textrm{illegal \& FP}}{\textrm{FP}}$ 
 & $\frac{\textrm{illegal}}{\textrm{total}}$ \\
\midrule
Resume & 1445  & 1 & 68 & 17 & 1.4\% & 20\% & 1.2\% \\
MSRA & 5853  & 6  & 318 & 107 & 1.9\% & 25\% & 1.8\% \\
Ontonotes & 5323  & 5  & 1336 & 314 & 1.6\% & 19\% & 4.6\%   \\
Weibo &  277 & 2  & 124   & 46 & 1.6\% & 27\% & 10.7\%  \\
ATIS  & 1643 & 0 & 70 & 24 & 0.0\% & 26\% &  1.4\% \\
SNIPS  & 1542 & 13 & 237 & 156 & 5.2\% & 40\% & 8.7\% \\
CoNLL2000  & 22957 & 36 & 888 & 100 & 3.9\% & 10\% & 0.6\% \\
CoNLL2003  & 5131 & 2 & 535 & 74 & 0.4\% & 12\% & 1.3\% \\
\bottomrule
\end{tabular}
\end{center}
\caption{ Statistics of the predicted text segments by category over a variety of sequence labeling datasets. A BERT-CRF model with BIO scheme is trained for each of the dataset, and the statistics are computed on the respective dev set.
 When the model generates an illegal path, we determine the predicted segments as in \citep{CONLL2000}, see Section  \ref{section_strategies} for more details. 
In the table ``TP'' and ``FP'' refer to ``True Positive'' and ``False Positive'' respectively. The column named ``$\frac{\textrm{illegal \& TP}}{\textrm{illegal}}$''
indicates the proportion of illegal segments that are correct predictions. The column named ``$\frac{\textrm{illegal \& FP}}{\textrm{FP}}$'' indicates the proportion of erroneous predictions that are due to illegal segments. The column named ``$\frac{\textrm{illegal}}{\textrm{total}}$''  stands for the proportion of illegal segments over all predictions.
\label{illegal_proportion}}
\end{table*}

The main contribution of this paper is to give a principled solution to the illegal path problem. More precisely:
\begin{enumerate}
\setlength\itemsep{0em}

\item We show that in the neural-CRF framework the illegal path problem is intrinsic and may accounts for non-negligible proportion (up to 40\%) of total errors. To the best of our knowledge we are the first to conduct this kind of study.
\item We propose Masked Conditional Random Field (MCRF), an improved version of the CRF that is by design immune to the illegal paths problem.
We also devise an algorithm for MCRF that requires only a few lines of code to implement.
\item We show in comprehensive experiments that MCRF performs significantly better than its CRF counterpart, and that its performance is on par with or better than more sophisticated models. Moreover, 
we achieve new State-of-the-Arts in two Chinese NER datasets.
\end{enumerate}

The remainder of the paper is organized as follows. Section 2 describes the illegal path problem and existing strategies that resolve it. In Section 3 we propose MCRF, its motivation and an approximate implementation. Section 4 is devoted to numerical experiments. We conclude the current work in Section 5.

\section{The illegal path problem}
\label{problem}
\subsection{Problem Statement}
As a common practice, most sequence labeling models utilize a certain tag encoding scheme to distinguish the boundary and the type of the text segments of interest.
 An encoding scheme makes it possible by introducing a set of tag prefixes and a set of tag transition rules.
For instance, the popular BIO scheme distinguishes the {\bf B}eginning, the {\bf I}nside and the {\bf O}utside of the chunks of interest, imposing that any {\tt I-$*$} tag must be preceded by a {\tt B-$*$} tag or another {\tt I-$*$} tag of the same type. 
Thus ``{\tt O O O I-LOC I-LOC O}'' is a forbidden sequence of tags because the transition {\tt O} $\to$ {\tt I-LOC} directly violates the 
BIO scheme design. 
Hereafter we shall refer to a sequence of tags that contains at least one illegal transition an \emph{illegal path}.

As another example, the BIOES scheme further identifies the {\bf E}nding of the text segments and the {\bf S}ingleton segments, thereby introducing more transition restrictions than BIO. e.g. an {\tt I-$*$} tag must always be followed by an {\tt E-$*$} tag of the same type,
 and an {\tt S-$*$} tag can only be preceded by an {\tt O}, an {\tt E-$*$} or another {\tt S-$*$} tag, etc. For a comparison of the performance of the encoding schemes, we refer to \citep{RATINOV2009} and references therein. 

When training a sequence labeling model with an encoding scheme, generally it is our hope that the model should be able to learn the semantics and the transition rules of the tags from the training data. However, even if the dataset is noiseless, a properly trained model may still occasionally make predictions that contains illegal transitions. This is especially the case for the CRF-based models, as there is no hard mechanism built-in to enforce those rules. The CRF ingredient by itself is only a \emph{soft} mechanism that encourages legal transitions and penalizes illegal ones. 

The hard transition rules might be violated when the model deems it necessary. To see this, let us consider a toy corpus where every occurrence of the token ``America'' is within the context of ``North America'', thus the token is always labeled as {\tt I-LOC}. Then, during training, the model may well establish the rule ``America $\Rightarrow$ {\tt I-LOC}'' (Rule 1), among many other rules such as ``an {\tt I-LOC} tag does not follow an {\tt O} tag'' (Rule 2), etc. Now consider the test sample ``Nathan left America last month'', which contains a stand-alone ``America'' labeled as {\tt B-LOC}. During inference, as the model never saw a stand-alone ``America'' before, it must generalize. If the model is more confident on Rule 1 than Rule 2, then it may yield an illegal output ``{\tt O O I-LOC O O}''.

\subsection{Strategies \label{section_strategies}}
The phenomenon of illegal path has already been noticed, but somehow regarded as trivial matters. For the BIO format, \citet{CONLL2000} have stated that
\begin{quotation}
The output of a chunk recognizer may contain inconsistencies in the chunk tags in case a word tagged {\tt I-X} follows a word tagged {\tt O} or {\tt I-Y}, with {\tt X} and {\tt Y} being different. These inconsistencies can be resolved by assuming that such {\tt I-X} tags starts a new chunk.
\end{quotation}
This simple strategy  has been adopted by CoNLL-2000 as a standard post-processing procedure\footnote{We are referring to the {\tt conlleval} script, available from \url{https://www.clips.uantwerpen.be/conll2000/chunking/}.} for the evaluation of the models' performance, and gain its popularity ever since.

We argue that such treatment is not only arbitrary, but also suboptimal. In preliminary experiments
we have studied the impact of the illegal path problem  using the BERT-CRF model for a number of tasks and datasets.
Our findings (see Table \ref{illegal_proportion}) suggest
that although the illegal segments only account for a small fraction (typically around 1\%) of total predicted segments, they constitute approximately a quarter of the false positives. Moreover, we found that only a few illegal segments are actually true positives. 
This raises the question of whether retaining the illegal segments is beneficial.
As a matter of fact, as we will subsequently show, 
a much higher macro F1-score can be obtained if we simply discard every illegal segments. 

\begin{figure*}[ht!]
\centering
\begin{tikzpicture}
[cnode/.style={draw=black, ,fill=#1, thick, minimum size=4mm, circle},  >=stealth, scale=0.75]

   
    \foreach \x in {1,...,9}
    {   \foreach \y in {3,...,7}
        {   
        \node[cnode=yellow!30] (s-\x\y) at (\x*1.5, \y*0.8) {};
        }
    }

\node[cnode=blue!30] (r-1) at (1*1.5, 7*0.8) {};
\node[cnode=blue!30] (r-2) at (2*1.5, 4*0.8) {};
\node[cnode=blue!30] (r-3) at (3*1.5, 7*0.8) {};
\node[cnode=blue!30] (r-4) at (4*1.5, 7*0.8) {};
\node[cnode=blue!30] (r-5) at (5*1.5, 7*0.8) {};
\node[cnode=blue!30] (r-6) at (6*1.5, 4*0.8) {};
\node[cnode=blue!30] (r-7) at (7*1.5, 5*0.8) {};
\node[cnode=blue!30] (r-8) at (8*1.5, 5*0.8) {};
\node[cnode=blue!30] (r-9) at (9*1.5, 7*0.8) {};

\node[cnode=green!30] (r-7) at (7*1.5, 3*0.8) {};
\node[cnode=green!30] (r-8) at (8*1.5, 3*0.8) {};

\node[left, brown] at (1, 7*0.8) {{\tt  \small O}};
\node[left, brown] at (1, 6*0.8) {{\tt  \small B-ORG}};
\node[left, brown] at (1, 5*0.8) {{\tt  \small I-ORG}};
\node[left, brown] at (1, 4*0.8) {{\tt  \small B-MISC}};
\node[left, brown] at (1, 3*0.8) {{\tt  \small I-MISC}};

\foreach \x in {1,...,10}
{
	\fill [blue!30, draw=black, rounded corners] (\x*1.5 - 0.7, 1.5 + 0.4) rectangle (\x*1.5 + 0.7, 1.5 - 0.2);
}

\foreach \x in {1,...,10}
{
	\fill [green!30, draw=black, rounded corners] (\x*1.5 - 0.7, .7 + 0.4) rectangle (\x*1.5 + 0.7, .7 - 0.2);
}

\foreach \x in {1,...,10}
{
	\fill [gray!30, draw=black, rounded corners] (\x*1.5 - 0.7, -.1 + 0.4) rectangle (\x*1.5 + 0.7, -.1 - 0.2);
}

\node[anchor=base, left] at (0.6, 1.6) {\small CRF prediction:};

\node[anchor=base] at (1*1.5, 1.5) {\scriptsize O};
\node[anchor=base] at (2*1.5, 1.5) {\scriptsize B-MISC};
\node[anchor=base] at (3*1.5, 1.5) {\scriptsize O};
\node[anchor=base] at (4*1.5, 1.5) {\scriptsize O};
\node[anchor=base] at (5*1.5, 1.5) {\scriptsize O};
\node[anchor=base] at (6*1.5, 1.5) {\scriptsize B-MISC};
\node[anchor=base] at (7*1.5, 1.5) {\scriptsize I-ORG};
\node[anchor=base] at (8*1.5, 1.5) {\scriptsize I-ORG};
\node[anchor=base] at (9*1.5, 1.5) {\scriptsize O};
\node[anchor=base] at (10*1.5, 1.5) {\scriptsize ...};

\node[anchor=base, left] at (0.6, 1) {\small MCRF prediction:};
\node[anchor=base, left] at (0.4, 0.6) {\small (Ground Truth)};

\node[anchor=base] at (1*1.5, 0.7) {\scriptsize O};
\node[anchor=base] at (2*1.5, 0.7) {\scriptsize B-MISC};
\node[anchor=base] at (3*1.5, 0.7) {\scriptsize O};
\node[anchor=base] at (4*1.5, 0.7) {\scriptsize O};
\node[anchor=base] at (5*1.5, 0.7) {\scriptsize O};
\node[anchor=base] at (6*1.5, 0.7) {\scriptsize B-MISC};
\node[anchor=base] at (7*1.5, 0.7) {\scriptsize I-MISC};
\node[anchor=base] at (8*1.5, 0.7) {\scriptsize I-MISC};
\node[anchor=base] at (9*1.5, 0.7) {\scriptsize O};
\node[anchor=base] at (10*1.5, 0.7) {\scriptsize ...};

\node[anchor=base, left] at (0.6, 0) {\small Input Tokens:};

\node[anchor=base] at (1*1.5, -.1) {\scriptsize The};
\node[anchor=base] at (2*1.5, -.1) {\scriptsize Briton};
\node[anchor=base] at (3*1.5, -.1) {\scriptsize who};
\node[anchor=base] at (4*1.5, -.1) {\scriptsize lost};
\node[anchor=base] at (5*1.5, -.1) {\scriptsize his};
\node[anchor=base] at (6*1.5, -.1) {\scriptsize World};
\node[anchor=base] at (7*1.5, -.1) {\scriptsize Boxing};
\node[anchor=base] at (8*1.5, -.1) {\scriptsize Council};
\node[anchor=base] at (9*1.5, -.1) {\scriptsize title};
\node[anchor=base] at (10*1.5, -.1) {\scriptsize ...};

\draw[->, thick] (s-17) -- (s-24);
\draw[->, thick] (s-24) -- (s-37);
\draw[->, thick] (s-37) -- (s-47);
\draw[->, thick] (s-47) -- (s-57);
\draw[->, thick] (s-57) -- (s-64);
\draw[->, thick, draw=black, dashed] (s-64) -- (s-75);
\draw[->, thick, draw=black] (s-75) -- (s-85);
\draw[->, thick, draw=black] (s-85) -- (s-97);

\draw[->, thick, draw=red] (s-64) -- (s-73);
\draw[->, thick, draw=red] (s-73) -- (s-83);
\draw[->, thick, draw=red] (s-83) -- (s-97);



\end{tikzpicture}
\caption{An example of CRF decoded path vs. MCRF decoded path. The CRF decoded path is represented as black arrows in the figure. This path contains one illegal transition (black dashed arrow) {\tt B-MISC}$\to${\tt I-ORG}, which results in two erroneous predictions: {\tt MISC} for ``World'' and {\tt ORG} for ``Boxing Council''. 
When using MCRF instead, the decoding algorithm has to search for an alternative path (red arrows), as all illegal transitions are blocked.
In this example, MCRF correctly predicts {\tt MISC} for the entity ``World Boxing Council''.
\label{fig_paths}
}
\end{figure*}
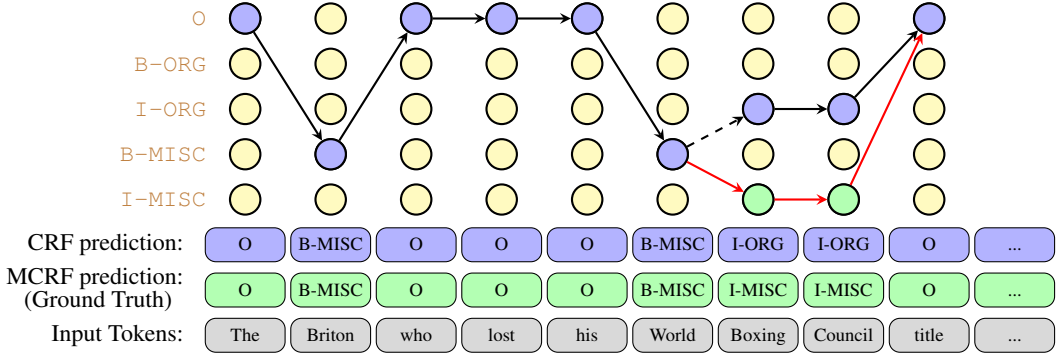

Although the strategy of discarding the illegal segments may be superior to that of \citep{CONLL2000}, it is nonetheless a hand-crafted, crude rule that lacks some flexibility.
To see this, let us take the example in Fig. \ref{fig_paths}. The prediction for text segment {\tt \small World Boxing Council} is ({\tt B-MISC}, {\tt I-ORG}, {\tt I-ORG}), which contains an illegal transition {\tt B-MISC}$\to${\tt I-ORG}. Clearly, neither of the post-processing strategies discussed above is capable of resolving the problem. 
Ideally, an optimal solution should convert the predicted tags to either ({\tt B-MISC}, {\tt I-MISC}, {\tt I-MISC}) or ({\tt B-ORG}, {\tt I-ORG}, {\tt I-ORG}), whichever is more likely. This is exactly the starting point of MCRF, which we introduce in the next section.

\section{Approach}

In this section we introduce the motivation and implementation of MCRF. We first go over the conventional neural-based CRF models in Section \ref{SECTION_CRF}. We then introduce  MCRF in Section \ref{SECTION_MCRF}. Its implementation will be given in Section \ref{SECTION_ALGORITHM}.

\subsection{Neural CRF Models\label{SECTION_CRF}}
Conventional neural CRF models typically consist of a neural network and a CRF layer. The neural network component serves as an encoder that usually first maps the input sequence of tokens 
  to a sequence of token encodings, which is then transformed (e.g. via a linear layer) into
 a sequence of token \emph{logits}. Each logit therein models the emission scores of the underlying token. 
The CRF component introduces a transition matrix that models the transition score from tag $i$ to tag $j$ for any two consecutive tokens.  By aggregating the emission scores and the transition scores, deep CRF models assign a score for each possible sequence of tags.

Before going any further, let us introduce some notations first. 
In the sequel, we denote by $x=\{x_1, x_2, \ldots, x_T\}$ a sequence of input tokens, by $y=\{y_1, \ldots, y_T\}$ their ground truth tags and by $l=\{l_1, \ldots, l_T\}$ the logits generated by the encoder network of the model.
Let $d$ be the number of distinct tags and 
denote by $[d] := \{1, \ldots, d\}$ the set of tag indices. Then $y_i\in[d]$ and $l_i\in\Rb^d$ for $1\leq i\leq T$.  
We denote by $W$ the set of all trainable weights in the encoder network, and by
$A=(a_{ij})\in\Rb^{d\times d}$ the transition matrix introduced by the CRF, where $a_{ij}$ is the transition score from tag $i$ to tag $j$. 
For convenience we call a sequence of tags a \emph{path}. 
For given input $x$, encoder weights $W$ and transition matrix $A$,  
we define the score of a path $p=\{n_1, \ldots, n_T \}$as
\ben
	s(p, x, W, A) = \sum_{i=1}^{T} l_{i, n_i} + \sum_{i=1}^{T-1} a_{n_i, n_{i+1}} ,
	\label{path}
	\label{score_of_path}
\een
where $l_{i, j}$ denotes the $j$-th entry of $l_i$. 
Let $\Sc$ be the set of all training samples, and $\Pc$ be the set of all possible paths. Then the loss function of neural CRF model is the average of negative log-likelihood over $\Sc$:
\begin{eqnarray}
\Lc(W, A) = - \frac{1}{|\Sc|}\!\!  \sum_{(x, y)\in\Sc} \!\!\!\log 	\frac{\exp{s(y, x)}}{\sum_{p\in\Pc} \exp{s(p, x)}} 
\label{crfloss}
\end{eqnarray}
where we have omitted the dependence of $s(\cdot, \cdot)$ on $(W, A)$ for conciseness.
One can easily minimize $\Lc(W, A)$ using any popular first-order methods such as SGD or Adam.

 Let $(W_{\textrm{opt}}, A_{\textrm{opt}})$ be a minimizer of $\Lc$. 
During decoding phase, the predicted path for a test sample $x_{\textrm{test}}$ is the path having the highest score, i.e.
\begin{eqnarray}
y_{\textrm{opt}} = \argmax{p\in\Pc} s(p, x_{\textrm{test}}, W_{\textrm{opt}}, A_{\textrm{opt}}).
\label{decoding}
\end{eqnarray}
The decoding problem can be efficiently solved by the Viterbi algorithm.

\subsection{Masked CRF\label{SECTION_MCRF}}
Our major concern on conventional neural CRF models is that no hard mechanism exists to enforce the transition rule, resulting in occasional occurrence of illegal predictions.

Our solution to this problem is very simple. 
Denote by $\Ic$ the set of all illegal paths.
We propose to constrain the ``path space'' in the CRF model to the space of all legal paths $\Pc/\Ic$, instead of the entire space of all possible paths $\Pc$. 
To this end, 
\begin{enumerate}
\setlength\itemsep{0em}
\item during training, the normalization term in  (\ref{crfloss}) should be the sum of the exponential scores of the legal paths;
\item during decoding, the optimal path should be searched over the space of all legal paths.
\end{enumerate}

The first modification above leads to the following new loss function:
\begin{eqnarray}
\nonumber
\Lc'(W, A) \!\!\!\!&:=& \!\!\!\!  - \frac{1}{|\Sc|}\!\!  \sum_{(x, y)\in\Sc} \!\!\!\log 	\frac{\exp{s(y, x)}}{\sum_{p\in\Pc / \Ic} \exp{s(p, x)}},\\
&& \!\!\!\!
\label{mcrfloss}
\end{eqnarray}
which is obtained by replacing the $\Pc$ in  (\ref{crfloss}) by $\Pc / \Ic$.

Similarly, the second modification leads to
\ben
y'_{\mathrm{opt}} = \argmax{p\in\Pc / \Ic}
 s(p, x_{\mathrm{test}}, W_{\mathrm{opt}}', A_{\mathrm{opt}}')
\label{decoding_mcrf}
\een
obtained by replacing the $\Pc$ in  (\ref{decoding}) by $\Pc / \Ic$,
where $(W_{\textrm{opt}}', A_{\textrm{opt}}')$ is a minimizer of (\ref{mcrfloss}). 

Note that the decoding objective (\ref{decoding_mcrf}) alone is enough to guarantee the complete elimination of illegal paths. However, this would create a mismatch between the training and the inference, as the model would attribute non-zero probability mass to the ensemble of the illegal paths. 
In Section \ref{main_result_section}, we will see that a naive solution based on (\ref{decoding_mcrf}) alone leads to suboptimal performance compared to a proper solution based on both (\ref{mcrfloss}) and (\ref{decoding_mcrf}).

\subsection{Algorithm \label{SECTION_ALGORITHM}}
\begin{figure}[ht!]
\centering
\begin{tikzpicture}
[cnode/.style={draw=black, ,fill=#1, thick, minimum size=7mm, rectangle},  >=stealth, scale=0.9]

  \foreach \x in {1,...,7}
    {   \foreach \y in {1,...,7}
        {   
        \node[cnode=yellow!30] (s-\x\y) at (\x, 8-\y) {$a_{\y\x}$};
        }
    }

\node[cnode=red!30] (m-31) at (3, 8-1) {$a_{13}$};
\node[cnode=red!30] (m-51) at (5, 8-1) {$a_{15}$};
\node[cnode=red!30] (m-71) at (7, 8-1) {$a_{17}$};

\node[cnode=red!30] (m-52) at (5, 8-2) {$a_{25}$};
\node[cnode=red!30] (m-72) at (7, 8-2) {$a_{27}$};

\node[cnode=red!30] (m-53) at (5, 8-3) {$a_{35}$};
\node[cnode=red!30] (m-73) at (7, 8-3) {$a_{37}$};

\node[cnode=red!30] (m-34) at (3, 8-4) {$a_{43}$};
\node[cnode=red!30] (m-74) at (7, 8-4) {$a_{47}$};

\node[cnode=red!30] (m-35) at (3, 8-5) {$a_{53}$};
\node[cnode=red!30] (m-75) at (7, 8-5) {$a_{57}$};

\node[cnode=red!30] (m-36) at (3, 8-6) {$a_{63}$};
\node[cnode=red!30] (m-56) at (5, 8-6) {$a_{65}$};

\node[cnode=red!30] (m-37) at (3, 8-7) {$a_{73}$};
\node[cnode=red!30] (m-57) at (5, 8-7) {$a_{75}$};

\node[anchor=base] at (1, 7.5) {\scriptsize O};
\node[anchor=base] at (2, 7.5) {\scriptsize B-LOC};
\node[anchor=base] at (3, 7.5) {\scriptsize I-LOC};
\node[anchor=base] at (4, 7.5) {\scriptsize B-ORG};
\node[anchor=base] at (5, 7.5) {\scriptsize I-ORG};
\node[anchor=base] at (6, 7.5) {\scriptsize B-PER};
\node[anchor=base] at (7, 7.5) {\scriptsize I-PER};


\node[left] at (0.6, 7) {\scriptsize O};
\node[left] at (0.6, 6) {\scriptsize B-LOC};
\node[left] at (0.6, 5) {\scriptsize I-LOC};
\node[left] at (0.6, 4) {\scriptsize B-ORG};
\node[left] at (0.6, 3) {\scriptsize I-ORG};
\node[left] at (0.6, 2) {\scriptsize B-PER};
\node[left] at (0.6, 1) {\scriptsize I-PER};



\end{tikzpicture}
\caption{An example of the masked transition matrix under the BIO scheme. 
 The entries in the red cells are masked as they correspond to illegal transitions. Under the BIO scheme, there are two types of illegal transitions: {\tt O} $\to$ {\tt I-X} for any {\tt X} and {\tt B-X} $\to$ {\tt I-Y} for any {\tt X}, {\tt Y} such that {\tt X} $\neq$ {\tt Y}.
\label{masked_transition_matrix}}
\end{figure}

Although in principle it is possible to directly minimize (\ref{mcrfloss}), thanks to the following proposition we can also achieve this via reusing the existing tools originally designed for minimizing (\ref{crfloss}), thereby saving us from making extra engineering efforts.
\begin{prop} \label{main_prop}
Denote by $\Omega\subset [d]\times[d]$ the set of all illegal transitions. For a given transition matrix $\A$, we denote by
 $\bar{A}(c)=\big(\bar{a}_{ij}(c)\big)$ the \emph{masked transition matrix} of $A$ defined as (see Fig. \ref{masked_transition_matrix})
\begin{eqnarray}
\bar{a}_{ij}(c) = \left\{ 
\begin{array}{ll}
c      & \textrm{if } (i, j)\in\Omega,  \\
a_{ij} & \textrm{otherwise},
\end{array}   \right.  \label{mask_transition}
\end{eqnarray}
where $c\ll 0$ is the \emph{transition mask}.
Then for arbitrary model weights $(W_0, A_0)$, we have
\ben
  \lim_{c\to - \infty} \Lc(W_0, \bar{A}_0(c)) & = & \Lc'(W_0, A_0)	\label{lim1}  \\
\!\!\!\!\!\!\!\!\! \lim_{c\to - \infty}\!\! \nabla_W \Lc(W_0, \bar{A}_0(c)) & = & \nabla_W \Lc'(W_0, A_0) \label{lim2} 
\een
and for all $(i,j)\in\Omega$
\ben
\!\!\!\!\!\!\!\!\!\!\!\!  \lim_{c\to - \infty}\!\! \nabla_{a_{ij}} \Lc(W_0, \bar{A}_0(c)) = \nabla_{a_{ij}} \Lc'(W_0, A_0). \label{lim3} 
\een
Moreover, for negatively large enough $c$ we have
\be
\argmax{p\in\Pc} s(p, x_{\textrm{test}}, W, A) = \argmax{p\in\Pc/\Ic} s(p, x_{\textrm{test}}, W, A)
\ee
\end{prop}
{\bf Proof. } See Appendix.

Proposition \ref{main_prop} states that for any given model state $(W, A)$, if we \emph{mask} the entries of $A$ that correspond to illegal transitions  (see Figure \ref{masked_transition_matrix}) by a negatively large enough constant $c$, then the two objectives (\ref{crfloss}) and (\ref{mcrfloss}), as well as their gradients, can be arbitrarily close. 
This suggests that the task of minimizing (\ref{mcrfloss})  can be achieved via minimizing (\ref{crfloss}) combined with
 keeping $A$ masked (i.e. making $a_{ij}=c$ constant for all $(i,j)\in \Omega$) throughout the optimization process. 

Intuitively, the purpose of transition masking is to penalize the illegal transitions in such a way that
 they will never be selected
during the Viterbi decoding, and the illegal paths as a whole only constitutes negligible probability mass during training. 

Based on Proposition \ref{main_prop}, we propose the \emph{Masked CRF} approach, formally described in Algorithm 1.
\begin{algorithm}[h!]
\caption{(MCRF) }
\begin{algorithmic}[1]
\State \textbf{Input:} Library for computing the gradients of conventional CRF loss (\ref{crfloss}),  training dataset $\Sc$, stopping criterion $\Cc$, set of illegal transitions $\Omega$, masking constant $c\ll 0$.
\State \textbf{Initialize:} model weight $W$ and tag transition matrix $A=(a_{ij})$.
\While{$\Cc$ is not met}
\State Sample a mini-batch from $\Sc$
\State Update $W$ and $A$ based on batch gradient
\For{$(i,j)\in\Omega$}
\State {$a_{ij} \leftarrow c$} \Comment{{\color{blue}maintain the mask}}
\EndFor
\EndWhile    
\State \textbf{Output: } Optimized $W$ and $A$.
\end{algorithmic}
\end{algorithm}

\section{Experiments}
In this section, we run a series of experiments\footnote{Our code is available on \url{https://github.com/DandyQi/MaskedCRF}.} to evaluate the performance of MCRF.
The datasets used in our experiments are listed as follows:
\begin{itemize}
\setlength\itemsep{0em}
\item {\bf Chinese NER:} OntoNotes 4.0 \citep{ONTONOTES_4}, MSRA \citep{MSRA}, Weibo \citep{WEIBO} and Resume \citep{LATTICE}. 
\item {\bf English NER:} CoNLL2003 \citep{CONLL2003}
\item {\bf Slot Filling:} ATIS \citep{ATIS} and SNIPS \citep{SNIPS}
\item {\bf Chunking:} CoNLL2000 \citep{CONLL2000}
\end{itemize}
The statistics of these datasets are summarized in Table \ref{dataset_statistics}. \begin{table}[h!]
\resizebox{\columnwidth}{!}{%
\begin{tabular}{c|cccccc}
\toprule 
dataset & task &  lan. & labels &  train & dev & test \\
\midrule
{Resume} & { NER} & CN & 8  &  3.8k & 472 & 477  \\
{MSRA} & { NER}  & CN  & 3 & 46.3k & - & 4.3k \\
{ Ontonotes} & { NER}  & CN  & 4 & 15.7k& 4.3k & 4.3k   \\
{ Weibo} &  { NER} & CN  & 7   & 1.3k & 270 & 270  \\
\midrule
{ ATIS}  & { SF} & EN & 79 & 4.5k & 500 & 893  \\
{ SNIPS}  & { SF} & EN & 39 & 13.0k & 700 & 700  \\
\midrule
{ CoNLL2000}  & { Chunk.} & EN & 11 & 8.9k & - & 2.0k  \\
\midrule
{ CoNLL2003}  & { NER} & EN & 4 & 14.0k & 3.2k & 3.5k  \\
\bottomrule
\end{tabular}
}
\caption{ \label{dataset_statistics} Statistics of the datasets.}
\end{table}

For Chinese NER tasks, we use the public-available\footnote{\url{https://github.com/google-research/bert}} $\textrm{BERT}_{\textrm{BASE}}$ as the pretrained model. For English NER and Chunking tasks, we use the cased version of $\textrm{BERT}_{\textrm{BASE}}$ model. We use uncased $\textrm{BERT}_{\textrm{BASE}}$ for English slot filling tasks. 

\begin{table*}[t]
\begin{center}
\begin{tabular}{l|cccc}
\toprule
 & {Resume} &  {MSRA} & {Ontonotes} & {Weibo} \\
\midrule 
\midrule
Lattice \citep{LATTICE}                   & 94.5        & 93.2          & 73.9           &  58.8       \\
Glyce \citep{GLYCE}$^\dagger$             & \underline{96.5}   & 95.5         & 81.6          & 67.6      \\
SoftLexicon \citep{SOFT_LEXICON}$^\dagger$      & 96.1 	     & 95.4          & 82.8    & \underline{70.5}      \\
FLAT \citep{FLAT}$^\dagger$              & 95.9               & 96.1         & 81.8                 &  68.6            \\
MRC \citep{MRC}$^\dagger$                & -                 & 95.7          & 82.1         &  -  \\
DSC \citep{DICE_LOSS}$^\dagger$ & - & \underline{96.7} & \underline{84.5} & - \\
\midrule
BERT-tagger-retain    & 95.7 {\small(94.7)} & 94.0 {\small(92.7)} & 78.1 {\small(76.8)} & 67.7 {\small(65.3)}  \\
BERT-tagger-discard   & 96.2 {\small(95.5)} & 94.6 {\small(93.6)} & 80.7 {\small(79.2)} & 69.7 {\small(67.5)}   \\
BERT-CRF-retain       & 95.9 {\small(94.8)} & 94.2 {\small(93.7)} & 81.8 {\small(81.2)} & 70.8 {\small(64.5)} \\
BERT-CRF-discard      & 97.2 {\small(96.6)} & 95.5 {\small(94.9)} & 83.1 {\small(82.4)} & 71.9 {\small(65.7)} \\
BERT-MCRF-decoding    & 97.3 {\small(96.6)} & 95.6 {\small(95.0)} & 83.2 {\small(82.5)} & 72.2 {\small(65.8)} \\ 
BERT-MCRF-training    & {\bf 97.6} {\small(96.9}) & {\bf 95.9} {\small(95.3)} & {\bf 83.7} {\small(82.7)} & {\bf 72.4} {\small(66.5)} 	\\  
\bottomrule
\end{tabular}
\end{center}
\caption{Results on Chinese NER datasets.
The ``$\dagger$'' symbol implies that the reported result is based on BERT.  The numbers in the parenthesis and the numbers next to it indicate the average and max F1-score, respectively.  
\label{main_result}}
\end{table*}

In preliminary experiments, we found out that the \emph{discriminative fine-tuning} approach \citep{ULMFIT} yields slightly better results than the standard fine-tuning  as recommended by \citep{BERT}. In discriminative fine-tuning, one uses different learning rates for each layer. 
Let $r_L$ be the learning rate for the last ($L$-th) layer and  $\eta$ be the decay factor. Then the learning rate for the $(L-n)$-th layer is given by $r_{L-n} = r_L \eta^{n}$. 
In our experiments, we use $r_{L}\in\{1e-4, 5e-5\}$ and $\eta\in\{1/2, 2/3\}$ depending on the dataset. The standard Adam optimizer is used throughout, and the mini-batch size is fixed to be 32. We always fine-tune for 5 epochs or 10000 iterations, whichever is longer. 

\subsection{Main results} \label{main_result_section}
In this section we present the MCRF results on 8 sequence labeling datasets. The baseline models are the following: 
\begin{itemize}
\setlength\itemsep{0em}
\item {\bf BERT-tagger:} The output of the final hidden representation for to each token is fed into a classification layer over the label set without using CRF. This is the approach recommended in \citep{BERT}. 
\item {\bf BERT-CRF:} BERT followed by a CRF layer, as is described in Section \ref{SECTION_CRF}. 
\end{itemize}
We use the following strategies to handle the illegal segments (See Table \ref{table_example} for an example):
\begin{itemize}
\setlength\itemsep{0em}
\item {\bf retain:} Keep and retag the illegal segments. This strategy agrees with \citep{CONLL2000}.
\item {\bf discard:} Discard the illegal segments completely. 
\end{itemize}
\begin{table}[h!]
\begin{tabular}{rccccc}
\toprule
original:  & {\tt O} & {\tt I-PER} & {\tt O}  & {\tt B-LOC} & {\tt I-MISC} \\
\midrule
retain:  & {\tt O} & {\tt B-PER} & {\tt O}  & {\tt B-LOC} & {\tt B-MISC} \\
discard: & {\tt O} & {\tt O}     & {\tt O}  & {\tt B-LOC} & {\tt O} \\
\bottomrule
\end{tabular}
\caption{An example illustrating the difference between ``retain'' strategy and ``discard'' strategy. 
\label{table_example}}
\end{table}
We distinguish two versions of MCRF:
\begin{itemize}
\setlength\itemsep{0em}
\item {\bf MCRF-decoding:} A naive version of MCRF that does masking only in decoding. The training process is the same as that in conventional CRF.
\item {\bf MCRF-training:} The proper MCRF approach proposed in this work. The masking is maintained in the training, as is described in Section \ref{SECTION_ALGORITHM}.
 We also refer to it as the MCRF for simplicity.
\end{itemize}

For each dataset and each model 
we ran the training 10 times with different random initializations and selected the model that performed best on the dev set for each run. We report the best and the average test F1-scores as the final results. If the dataset does not provide an official development set, we randomly split the training set and use 10\% of the samples as the dev set. 

\subsubsection{Results on Chinese NER \label{experiment_chinese_ner}}
The results on Chinese NER tasks are presented in Table \ref{main_result}. It can be seen that 
the MCRF-training approach significantly outperforms all baseline models and establishes new State-of-the-Arts for Resume and Weibo datasets. From these results we can assert that the improvement brought by MCRF is mainly due to the effect of masking in training, not in decoding.
Besides, we notice that the ``discard'' strategy substantially outperforms the ``retain'' strategy, which agrees with the statistics presented in Table \ref{illegal_proportion}.

\begin{figure}[h!]
\begin{center}
  \includegraphics[width=0.3\textwidth]{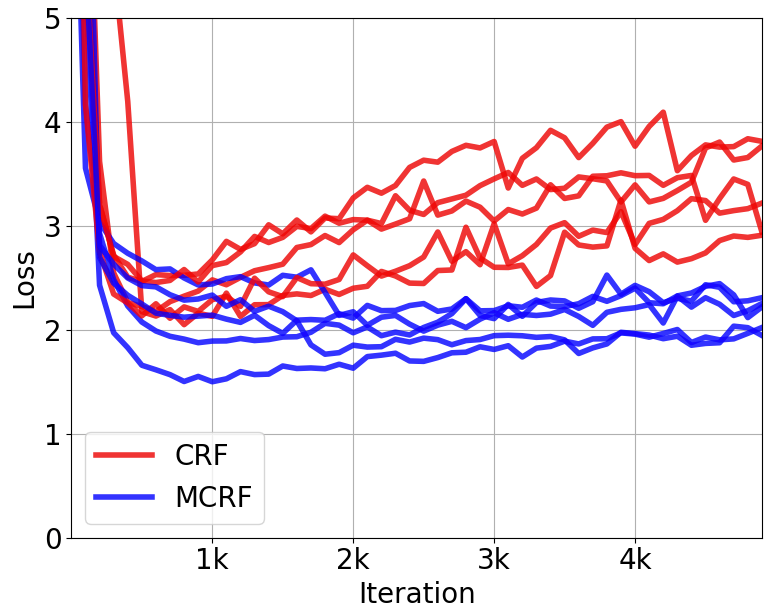}
 \end{center}
  \caption{Curves of dev loss for CRF and MCRF.  
\label{loss_curve}
 }
\end{figure}

We also plotted in Fig. \ref{loss_curve} the loss curves of CRF and MCRF on the development set of MSRA. It can be clearly seen that MCRF incurs a much lower loss during training.
This confirms our hypothesis that the CRF model attributes non-zero probability mass to the ensemble of the illegal paths, as otherwise
the denominators in (\ref{mcrfloss}) and in (\ref{crfloss}) would have been equal, and in that case the loss curves of CRF and MCRF would have converged to the same level.

Note that some of the results listed in Table \ref{main_result} are based on models that utilize additional resources.
\citet{LATTICE} and \citet{SOFT_LEXICON} utilized Chinese lexicon features to enrich the token representations.
\citet{GLYCE} combined Chinese glyph information with BERT pre-training.
In contrast, the proposed MCRF approach is simple yet performant. It achieves comparable or better results without relying on additional resources. 

\subsubsection{Results on Slot Filling}
One of the main features of the AITS and SNIPS datasets is the large number of slot labels (79 and 39 respectively) with relatively small training set (4.5k and 13k respectively). 
This requires the sequence labeling model learn the transition rules in a sample-efficient manner.
Both ATIS and SNIPS provide an intent label for each utterance in the datasets, but in our experiments we did not use this information and 
rely solely on the slot labels.

The results are reported in Table \ref{result_sf}. It can be seen that MCRF-training outperforms the baseline models and achieves competitive results compared to previous published results. 
\begin{table}[h!]
\begin{center}
\resizebox{1.0\columnwidth}{!}{%
\begin{tabular}{l|c|c}
\toprule
{\bf Model} & {ATIS} &  {SNIPS} \\
\midrule 
\citep{SLOT_GATED}  &  95.4 & 89.3 \\
\citep{ATIS_SOTA} & \underline{96.5} & - \\ 
\citep{SF_CAPSULE} & 95.2 & 91.8	 \\
\citep{e-etal-2019-novel}   & 95.8    & 92.2               \\
\citep{SIDDHANT2019}   & 95.6      & \underline{93.9}           \\
\midrule
BERT-tagger-retain     & 95.2 {\small(92.9)}         & 93.2 {\small(92.1)}        \\
BERT-tagger-discard    & 95.6 {\small(93.1)}         & 93.5 {\small(92.3)}       \\
BERT-CRF-retain        & 95.5 {\small(93.5)}         & 94.6 {\small(93.7)} 		  \\
BERT-CRF-discard       & 95.8 {\small(93.9)}         & 95.1 {\small(94.3)}       \\
BERT-MCRF-decoding     & 95.8 {\small(93.9)}         & 95.1 {\small(94.4)}       \\
BERT-MCRF-training     & {\bf 95.9} {\small(94.4)}   & {\bf 95.3} {\small(94.6)}    \\
\bottomrule
\end{tabular}
}
\end{center}
\caption{Test F1-scores on slot filling datasets. \label{result_sf}}
\end{table}

\subsubsection{Results on Chunking}
The results on CoNLL2000 chunking task are reported in Table. \ref{RES_CHUNKING}. The proposed MCRF-training outperforms the CRF baseline by 0.4 in F1-score.
\begin{table}[h!]
\begin{center}
\begin{tabular}{l|c}
\toprule
{\bf Model}  & F1  \\
\midrule 
ELMo \citep{PETERS2017}              & 96.4       \\
CSE \citep{AKBIK2018}               & 96.7       \\
GCDT \citep{GCDT} & \underline{97.3}       \\
\midrule
BERT-tagger-retain     & 96.1 {\small(95.7)}     \\
BERT-tagger-discard & 96.3 {\small(96.0)}     \\
BERT-CRF-retain  & 96.5 {\small(96.2)}   \\
BERT-CRF-discard & 96.6 {\small(96.3)}    \\
BERT-MCRF-decoding & 96.6 {\small(96.4)}    \\
BERT-MCRF-training & {\bf 96.9} {\small(96.5)}  \\
\bottomrule
\end{tabular}
\end{center}
\caption{Results on CoNLL2000 chunking task. \label{RES_CHUNKING}}
\end{table}


\subsection{Ablation Studies}
In this section, we investigate the influence of various factors that may impact the performance of MCRF.
In particular, we are interested in the quantity \emph{MCRF gain}, which we denote by $\Delta$, defined simple as the difference of F1-score of MCRF-training and that of the conventional CRF (with either ``retain'' or ``discard'' strategy). 

\subsubsection{Effect of Tagging Scheme}
In the previous experiments we have always used the BIO scheme. 
It is of interest to explore the performance of MCRF under other tagging schemes such as BIOES.
The BIOES scheme is considered more expressive than BIO as it introduces more labels and more transition restrictions. 

\begin{figure}[t]
\centering
  \includegraphics[width=0.45\textwidth]{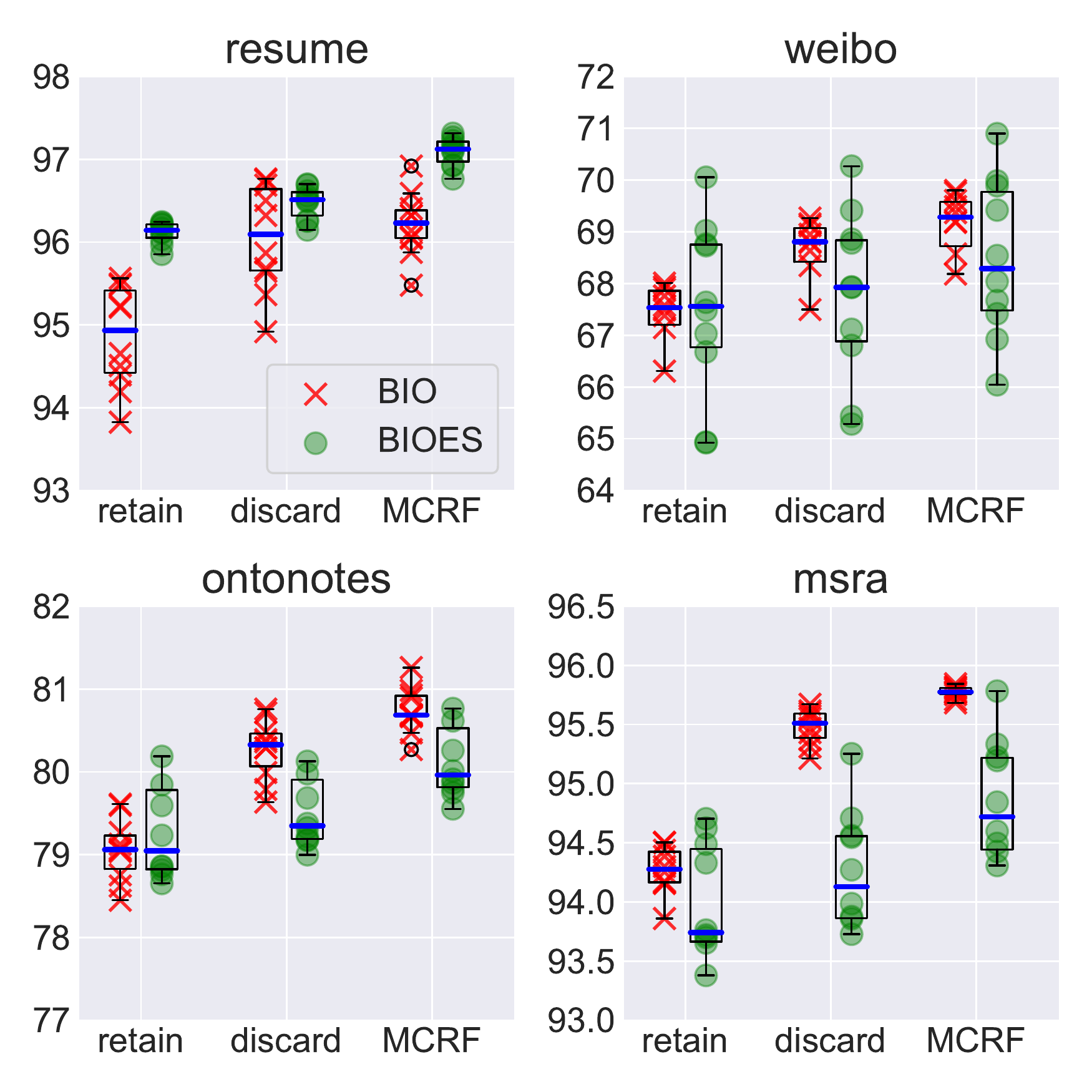}
  \caption{Ablation over the tagging scheme (BIO vs. BIOES).   The F1-scores on the dev sets are plotted. 
  \label{bio_vs_bioes}
 }
\end{figure}

We have re-run the experiments in Section \ref{experiment_chinese_ner} using the BIOES scheme.  Our results are reported in  Fig. \ref{bio_vs_bioes} and Table  \ref{mcrf_gain}. 
 It is clearly seen that under the BIOES scheme MCRF still always outperforms the CRF baselines.
 Note that compared to the case under BIO scheme, the MCRF gain is less significant against the CRF-retain baseline, but larger against CRF-discard.

\begin{table}[h!]
\begin{center}
\begin{tabular}{l|cc|cc}
\toprule
 & \multicolumn{2}{c|}{BIO}  & \multicolumn{2}{c}{BIOES} \\
\midrule
& $\Delta_{\textrm{ret.}}$ & $\Delta_{\textrm{disc.}}$  & $\Delta_{\textrm{ret.}}$  & $\Delta_{\textrm{disc.}}$  \\
Resume      & 2.1 & 0.3        & 1.0 & 0.6       \\
MSRA        & 1.6 & 0.4        & 0.8 & 0.6       \\
Ontonotes   & 1.5 & 0.3        & 0.9 & 0.6       \\
Weibo       & 2.0 & 0.8        & 0.9 & 0.8       \\
\bottomrule
\end{tabular}
\end{center}
\caption{A comparison of the average MCRF gain under BIO and BIOES schemes. 
The symbols $\Delta_{\textrm{ret.}}$ and $\Delta_{\textrm{disc.}}$ stand for the gain against BERT-retain and BERT-discard, respectively.
\label{mcrf_gain}}
\end{table}

\subsubsection{Effect of Sample Size}
One may hypothesize that the occurrence of illegal paths might be due to the scarcity of training data, i.e.  
a model should be less prone to illegal paths if the training dataset is larger.
To test this hypothesis, we randomly sample 10\% of the training data from MSRA and Ontonotes, creating a smaller version of the respective dataset.
We compare the proportion of the illegal segments produced by BERT-CRF trained on the original dataset with the one trained on the smaller dataset. 
We also report the performance gain brought by MCRF in these two scenarios. Our findings are summarized in Table \ref{small_sample_table}.  
As can be seen from the table, 
the models trained with fewer data do yield slightly more illegal segments, but
the MCRF gains under the two scenarios are close.
\begin{table}[h!]
\begin{center}
\resizebox{0.9\columnwidth}{!}{%
\begin{tabular}{l|ccc|ccc}
\toprule
& \multicolumn{3}{c|}{MSRA-full} &  \multicolumn{3}{c}{MSRA-10\%} \\
               & ill. & F1 & $\Delta$ & ill. & F1 & $\Delta$ \\
\midrule
retain	       & 1.8\%& 94.2 &  1.6      & 2.4\% & 90.4 & 1.2  \\
discard        & -    & 95.4 &  0.5      & -     & 90.7 & 0.9 \\
MCRF           & 0\%  & 95.8 & -         & 0\%   & 91.6 & -  \\
\bottomrule
\toprule
& \multicolumn{3}{c|}{Ontonotes-full} &  \multicolumn{3}{c}{Ontonotes-10\%} \\
               &ill. & F1 & $\Delta$ & ill. & F1 & $\Delta$ \\
\midrule
retain	       & 4.2\% & 79.2 &  1.6      & 4.7\% & 78.7 &  1.2 \\
discard        & -     & 80.4 &  0.4      & -     & 79.1 &  0.8  \\
MCRF           & 0\%   & 80.8 & -         & 0\%   & 79.9 &  -  \\
\bottomrule
\end{tabular}
}
\end{center}
\caption{
Ablation over the training set size.
The column named ``ill.'' indicates the proportion of illegal segments over all predicted segments.
 \label{small_sample_table}}
\end{table}

\subsubsection{Effect of Encoder Architecture}
So far we have experimented with BERT-based models. Now we explore effect of neural architecture. We trained a number of models on CoNLL2003 with varying encoder architectures. The key components are listed as follows:
\begin{itemize}
\setlength\itemsep{0em}
\item ELMo: pretrained language model\footnote{Model downloaded from \url{https://github.com/allenai/bilm-tf}} that serves as an sequence encoder.
\item CNN: CNN-based character embedding layer, with weights extracted from pretrained ELMo. It is used to generate word embeddings for arbitrary input tokens.
\item LSTM-$n$: $n$-layer bidirectional LSTM with hidden dimension $h=200$.
\end{itemize}
The results of our experiments are given in Table \ref{ablation_encoder}.
We observe that the encoder architecture has a large impact on the occurrence of illegal paths, and
the BERT-based models appear to generate much more illegal paths than ELMo-based ones. 
This is probably due to the fact that transformer-encoders are not sequential in nature.
A further study is needed to investigate this phenomenon, but it is beyond the scope of the current work.
We also notice that the MCRF gain seems to be positively correlated with the proportion of the illegal paths generated by the underlying model. This is expected, since
the transition-blocking mechanism of MCRF will (almost) not take effect if the most probable path estimated by the underlying CRF model is already legal. 


\begin{table}[h]
\begin{center}
\resizebox{\columnwidth}{!}{%
\begin{tabular}{l|ccccc}
\toprule
{\bf Encoder}      & ill. & err.    & CRF & MCRF & $\Delta$  \\
\midrule
LSTM-1 & 3.1\% & 11.7\%         & 82.2 & 83.2 & 1.0 \\
LSTM-2 & 1.4\% & 8.3\%          & 84.3 & 85.1 & 0.8 \\
CNN + LSTM-1 & 0.4\%  & 4.0\%   & 94.1 & 94.3 & 0.2 \\
CNN + LSTM-2 & 0.3\%  & 2.3\%   & 94.0 & 94.5 & 0.5 \\
ELMo + LSTM-1 & 0.4\% & 3.3\%   & 95.1 & 95.3 & 0.2\\
ELMo + LSTM-2 & 0.6\% & 5.5\%   & 95.0 & 95.3 & 0.3 \\
BERT          & 1.3\% & 12.5\%  & 94.5 & 95.4 & 0.9\\
BERT + LSTM-1 & 1.0\% & 13.1\%  & 94.7 & 95.3 & 0.6\\
BERT + LSTM-2 & 0.9\% & 10.3\%  & 93.9 & 95.0 & 1.1\\
\bottomrule
\end{tabular}
}
\end{center}
\caption{Ablation over the encoder models. 
The column named ``err.'' indicates the proportion of erroneous predictions that are due to illegal segments.
\label{ablation_encoder}}
\end{table}

\subsection{Related Work}
Some models are able to solve sequence labeling tasks without relying on BIO/BIOES type of tagging scheme to distinguish the boundary and the type of the text segments of interest, thus do not suffer from the illegal path problems. For instance,
Semi-Markov CRF \citep{SARAWAGI2005} uses an additional loop to search for the segment spans, and directly yields a sequence of segments along with their type. The downside of Semi-Markov CRF is that it incurs a higher time complexity compared to the conventional CRF approach. Recently, \citet{MRC} proposed a Machine Learning Comprehension (MRC) framework to solve NER tasks. Their model uses two separate binary classifiers to predict whether each token is the start or end of an entity, and an additional module to determine which start and end tokens should be matched.

We notice that the CRF implemented in PyTorch-Struct \citep{TORCH_STRUCT} has a different interface than usual CRF libraries and it takes not two tensors for emission and transition scores, but rather one score tensor of the shape {(batch size, sentence length, number of tags, number of tags)}. This allows one to incorporate even more prior knowledge in the structured prediction by
setting a constraint mask as a function of not only a pair of tags, but also words on which the tags are assigned. Such feature may be exploited in future work.

Finally, we acknowledge that the naive version of MCRF that does constrained decoding has already been implemented in AllenNLP\footnote{\url{https://github.com/allenai/allennlp/blob/main/allennlp/modules/conditional_random_field.py}} \citep{ALLENNLP}. As shown in Section \ref{main_result_section}, such approach is suboptimal compared to the proposed MCRF-training method.

\section{Conclusion}
Our major contribution is the proposal of MCRF, a constrained variant of CRF that masks illegal transitions during CRF training, eliminating illegal outcomes in a principled way. 

We have justified MCRF from a theoretical perspective, and shown empirically in a number of datasets that MCRF consistently outperforms the conventional CRF.
As MCRF is easy to implement and incurs zero additional overhead, we advocate always using MCRF instead of CRF when applicable.

\section*{Acknowledgments}
We thank all anonymous reviewers for their valuable comments. We also thank Qin Bin and Wang Gang for their support. This work is also supported by the National
Natural Science Foundation of China (NSFC No. 61701547).

\newpage
\bibliographystyle{acl_natbib}
\bibliography{nlp,emnlp2020}

\newpage
\appendix
\section{Appendices}
\subsection{Proof of Proposition 1}

Denote by $L$ and $L'$ the likelihood function of  sample $(x, y)$ for CRF and MCRF model respectively:
\ben
L(W, A) & = & \frac{\exp{s(y, x, W, A)}}{\sum_{p\in\Pc} \exp{s(p, x, W, A)}}, \label{likelihood1} \\
L'(W, A) & = & \frac{\exp{s(y, x, W, A)}}{\sum_{p\in\Pc/\Ic} \exp{s(p, x, W, A)}} . \label{likelihood2}
\een
 To simplify the notations, we also write 
\be
L(W, A) = \frac{N(W, A)}{D(W, A)},\quad L'(W, A) = \frac{N(W, A)}{D'(W, A)},
\ee
where
\be
N(W, A) & = & \exp{s(y, x, W, A)} \\
D(W, A) & = & \sum_{p\in\Pc} \exp{s(p, x, W, A)} \\
D'(W, A) & = & \sum_{p\in\Pc/\Ic} \exp{s(p, x, W, A)}
\ee

Proposition 1 is a direct corollary of the following result:
\begin{lemm} \label{first_lemma}
Let $(x, y)$ be a sample with $y\in\Pc/\Ic$. Then
for arbitrary $(W_0, A_0)$, we have
\ben
  \lim_{c\to - \infty} L(W_0, \bar{A}_0(c)) & = & L'(W_0, A_0)	\label{lim1}  \\
\!\!\!\!\!\!\!\!\! \lim_{c\to - \infty}\!\! \nabla_W L(W_0, \bar{A}_0(c)) & = & \nabla_W L'(W_0, A_0) \label{lim2} 
\een
and for all $(i,j)\in\Omega$
\ben
\!\!\!\!\!\!\!\!\!\!\!\!  \lim_{c\to - \infty}\!\! \nabla_{a_{ij}} L(W_0, \bar{A}_0(c)) = \nabla_{a_{ij}} L'(W_0, A_0). \label{lim3} 
\een
\end{lemm}
{\bf Proof.} 
First, we recall that
 \ben
	s(p, x, W, A) = \sum_{i=1}^{T} l_{i, n_i} + \sum_{i=1}^{T-1} a_{n_i, n_{i+1}} ,
	\label{path}
	\label{score_of_path}
\een
and the masked transition matrix $\bar{A}(c)=\big(\bar{a}_{ij}(c)\big)$ is defined as 
\begin{eqnarray}
\bar{a}_{ij}(c) = \left\{ 
\begin{array}{ll}
c      & \textrm{if } (i, j)\in\Omega,  \\
a_{ij} & \textrm{otherwise},
\end{array}   \right.  \label{mask_transition}
\end{eqnarray}
where $\Omega$ is the set of illegal transitions. 

Since $\bar{A}(c)$ differs from $A$ only on entries corresponding to illegal transitions and a legal path contains only legal transitions, 
 it follows from (\ref{score_of_path})
 that $\forall p\in\Pc/\Ic$
\ben
s(p, x, W_0, \bar{A}_0(c)) = s(p, x, W_0, {A}_0).  \label{legalscore}
\een
Thus 
\ben
N(W_0, \bar{A}_0(c)) = N(W_0, A_0). \label{numerator}
\een
Next, we show
\ben
D'(W_0, \bar{A}_0(c)) \xrightarrow[c\to-\infty]{} D(W_0, A_0).  \label{denom}
\een
By (\ref{likelihood1}) (\ref{likelihood2}) and (\ref{legalscore}),  it suffices to demonstrate for any illegal path $p\in\Ic$
\ben \label{vanish}
\lim_{c\to-\infty}\exp s(p, x, W_0, \bar{A}_0(c)) = 0.
\een
To achieve this, we rewrite $s(p, x, W_0, \bar{A}_0(c))$ as a product of three terms:
\be
&& \exp s(p, x, W_0, \bar{A}_0(c)) \\
&& \quad = \prod_{i=1}^T e^{l_{i, n_i}}
\!\!\!\!\!\!\! \underbrace{\prod_{(i,j)\in \Tc/\Omega \atop (i,j)\sim p}}_{\text{legal transitions}} 
\!\!\!\!\!\!\! e^{\bar{a}_{ij}(c)}
\!\!\!\!\!\!\! \underbrace{\prod_{(i,j)\in\Omega \atop (i,j) \sim p}}_{\text{ illegal transitions}} 
\!\!\!\!\!\!\! e^ {\bar{a}_{ij}(c)}
\ee
where $(i,j)\sim p$ means that $(i,j)$ is a transition contained in path $p$. 
Let $E(p)$ be the number of illegal transitions in $p$.
If $p$ is illegal, then $E(p) >0$; otherwise $E(p)=0$. Since $\bar{a}_{ij}(c)=c$ for $(i,j)\in\Omega$ by definition, 
\be
&&\exp s(p, x, W_0, \bar{A}_0(c)) \\
&& \quad\quad = \Big(\prod_{i=1}^T e^{l_{i, n_i}} 
\prod_{(i,j)\in \Tc/\Omega \atop (i,j)\sim p} e^{a_{ij}}\Big)\cdot
e^ {c E(p)}.
\ee
Now that the terms in the parenthesis do not depend on $c$ and $e^{cE(p)}$ vanishes as $c\to -\infty$, we achieve (\ref{vanish}). Then (\ref{lim1}) of Lemma \ref{first_lemma} is proved.

Now we turn to the proof of (\ref{lim2}). By elementary calculus we have
\be
\nabla_W L & = & \Big({D\cdot \nabla_W N  - N\cdot \nabla_W D}\Big)\cdot{D^{-2}} \\
\nabla_W L' & = & \Big({D'\cdot \nabla_W N  - N\cdot \nabla_W D'}\Big)\cdot{D'^{-2}}.
\ee
By (\ref{numerator}) and  (\ref{denom}), it remains to show 
\ben\label{dprime}
\!\!\!\!\!\!\!\!\!\!\!\!\!\!  \nabla_W D'(W_0, \bar{A}_0(c)) \xrightarrow[c\to-\infty]{} \nabla_W D(W_0, A_0).
\een
By the same argument as in the proof of (\ref{legalscore}) and (\ref{vanish}), it is easily seen that for $p\in\Pc/\Ic$
\be
&& \nabla_W \Big( \exp s(p, x, W, A) \Big) \Big|_{W_0, A_0}\\
&& \quad = \nabla_W \Big( \exp s(p, x, W, A) \Big) \Big|_{W_0, \bar{A}_0(c)}
\ee
and for $p\in\Ic$
\be
\nabla_W \Big( \exp s(p, x, W, A) \Big) \Big|_{W_0, \bar{A}_0(c)} \xrightarrow[c\to-\infty]{} 0.
\ee
Thus (\ref{dprime}) is achieved and (\ref{lim2}) follows.

Finally, the proof of (\ref{lim3}) is similar to that of (\ref{lim2}). 

\end{document}